\documentclass[letterpaper]{article} 
\usepackage[preprint]{aaai2027}  
\usepackage[hyphens]{url}  
\usepackage{graphicx} 
\usepackage{natbib}  
\usepackage{caption} 
\usepackage{algorithm}
\usepackage{algorithmic}

\usepackage{newfloat}
\usepackage{listings}
\DeclareCaptionStyle{ruled}{labelfont=normalfont,labelsep=colon,strut=off} 
\floatstyle{ruled}
\newfloat{listing}{tb}{lst}{}
\floatname{listing}{Listing}

\usepackage{booktabs}
\newcommand{\flowdancewidefigure}[3]{%
    \begin{figure*}[t]
        \centering
        \includegraphics[width=\textwidth]{#1}
        \caption{#2}
        \label{#3}
    \end{figure*}%
}

\newcommand{\flowdancecolumnfigure}[3]{%
    \begin{figure}[t]
        \centering
        \includegraphics[width=\columnwidth]{#1}
        \caption{#2}
        \label{#3}
    \end{figure}%
}

\newcommand{\flowdanceteaserfigure}[3]{%
    \begin{center}
        \includegraphics[width=\textwidth]{#1}
        \captionof{figure}{#2}
        \label{#3}
    \end{center}%
}

\title{FlowDance: Music-Driven Dance Video Generation with Parallel Pose \\and RGB Streams}
\author {
    Genying Li\textsuperscript{\rm 1,\rm 2}\equalcontrib,
    Boda Lin\textsuperscript{\rm 1}\equalcontrib,
    Jiachen Li\textsuperscript{\rm 1},
    Zijian Jia\textsuperscript{\rm 1,\rm 2},
    Haojie Zheng\textsuperscript{\rm 2,\rm 3},
    Yiming Wang\textsuperscript{\rm 1},\\
    Shuchen Weng\textsuperscript{\rm 2}\corresponding,
    Si Li\textsuperscript{\rm 1}\corresponding
}
\affiliations {
    \textsuperscript{\rm 1}School of Artificial Intelligence, Beijing University of Posts and Telecommunications\\
    \textsuperscript{\rm 2}Beijing Academy of Artificial Intelligence\\
    \textsuperscript{\rm 3}School of Software and Microelectronics, Peking University\\
}

\begin{document}

\newcommand{\flowdanceteaser}{%
    \flowdanceteaserfigure
        {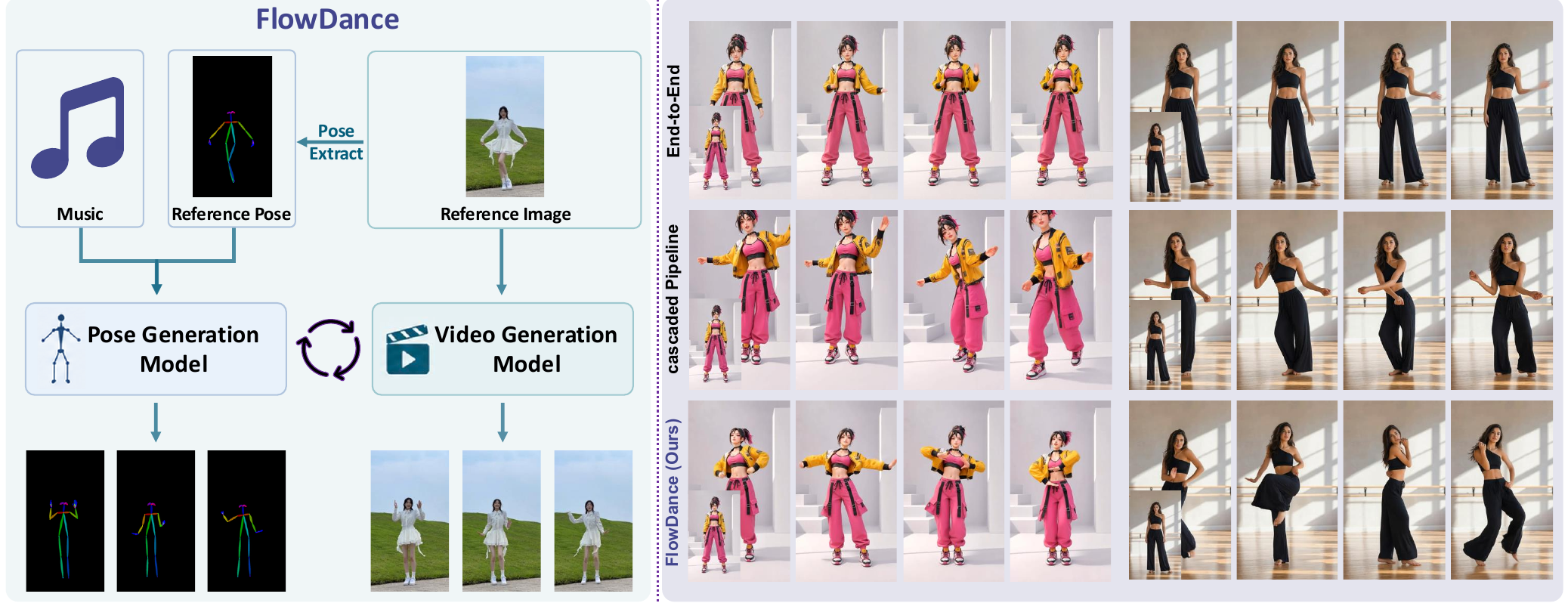}
        {Comparison with end-to-end and cascaded music-driven dance video synthesis. FlowDance generates pose and RGB videos in parallel, enabling continuous structural guidance without explicit 3D-to-2D pose conversion.}
        {fig:teaser}%
}

\makeatletter
\g@addto@macro\@maketitle{\flowdanceteaser}
\makeatother

\maketitle

\begin{abstract}
Music-driven dance video synthesis aims to animate a reference person according to a given music clip.
The task is challenging because it requires a model to jointly learn music-to-motion correspondence, identity-preserving human animation, temporal coherence, and visually realistic video generation.
We present FlowDance, a music-driven dance video generation framework that integrates explicit motion modeling with reference-preserving visual synthesis through parallel pose and RGB streams.
We further introduce timestep-aware pose injection to adapt structural guidance across denoising steps and persistent identity injection to preserve the reference appearance over long video.
To support this task, we further build a popularity-curated, high-resolution in-the-wild dance video dataset with synchronized music, RGB videos, 3D body motion, camera parameters, and projected 2D pose annotations. 
Extensive experiments show that FlowDance achieves strong performance in both dance motion generation and music-driven dance video synthesis.
The project page is available at https://ghost-love-you.github.io/Projects/FlowDance.
\end{abstract}


\section{Introduction}


Music and dance are deeply intertwined forms of human expression, where musical cues such as rhythm, melody, intensity, and style strongly influence the characteristics of dance movements.
This correspondence between music and motion has motivated a growing body of research on music-driven dance generation~\cite{li2021aist,siyao2022bailando,tseng2022edge,li2024lodge,zhang2025opendance}, where existing approaches primarily focus on synthesizing 3D human motion sequences conditioned on audio~\cite{loper2015smpl,pavlakos2019smplx}.
However, such 3D motion sequences only describe the underlying body dynamics and inevitably omit crucial visual factors such as dancer identity, appearance, and scene context, thereby limiting their direct applicability to real-world video content creation. Therefore, generating realistic dance videos in which specific individuals dance naturally in synchronization with music remains an important yet challenging problem.

Moving from dance motion generation to dance video generation introduces two additional challenges: preserving the identity and appearance of a reference person, including clothing, body shape, background, and fine visual details, and maintaining motion temporal coherence throughout the generated video. 
Thanks to advances in pose-guided human image animation, synthesizing a target person from a reference image and a 2D pose video has been extensively explored~\cite{zhou2019dance,hu2023animate,xu2023magicanimate,zhu2024champ,pang2024dreamdance}. 
This progress enables a prevalent decoupled pipeline for music-driven dance video generation~\cite{chen2025xdancer,wang2025choreomuse,yang2025macedance}: a music-to-motion model first predicts 3D body motion, often parameterized by SMPL or SMPL-X; a projector then maps these parameters into a 2D pose video, which drives an image animation model. 
Notably, by separating choreography from visual rendering, this formulation reduces the learning burden of each stage and has achieved promising generation quality. 
However, errors can accumulate across the pipeline, particularly during the projection from SMPL parameters to 2D poses, thereby propagating and amplifying inaccuracies downstream. 
How the projector should adapt its mapping to the target person's body shape, including stature, build, and body proportions, as well as the person's location in the reference image, remains insufficiently explored. 
The resulting pose misalignment may distort body structure, introduce visual artifacts, or move parts of the subject outside the video frame. 
Because the animation model only receives the projected pose sequence, it cannot correct these upstream errors, which consequently remain visible in the final synthesized output.

End-to-end methods instead map music and a reference image directly to RGB video~\cite{dong2025musedance}. 
They avoid explicit motion projection and allow choreography and rendering to be learned jointly, but require a single model to perform motion planning, identity preservation, temporal modeling, and photorealistic video synthesis in a high-dimensional visual space. 
This coupling makes optimization difficult and can weaken motion fidelity. These limitations suggest that motion should be modeled explicitly, but without reintroducing the cross-modal conversion and stage-wise isolation of cascaded pipelines.


In this paper, we propose \textbf{FlowDance}, a music-driven dance video generation framework with parallel pose and RGB streams.
As illustrated in Figure~\ref{fig:teaser}, given a reference image and a music track, FlowDance generates a music-aligned 2D pose video and a reference-preserving dance video in parallel.
By allowing the pose stream to continuously guide RGB stream throughout denoising, FlowDance integrates explicit motion modeling with reference-aware visual synthesis within a unified diffusion model.

We build our framework upon the pretrained Wan2.2~\cite{wan2025} video diffusion model to leverage its strong spatio-temporal priors and native video generation capabilities.
To eliminate explicit SMPL-to-2D projection and its associated representation gap, we introduce a pose stream that directly generates a DWPose-style pose video.
Specifically, the pose stream utilizes the pose extracted from the reference image as the initial frame of the pose sequence, anchoring subsequent motion generation to the target subject's body shape and image-space location.
A dedicated audio cross-attention module further condition the pose stream on the input music, enabling the generated motion to follow its rhythmic and stylistic cues.
Alongside the pose stream, a parallel RGB stream progressively denoises the target dance video using intermediate features from the corresponding pose-stream blocks as structural guidance.
This block-wise interaction enables motion generation and visual synthesis to be jointly refined throughout denoising, allowing the RGB stream to receive progressively updated motion guidance, avoiding the stage-wise handoff and error accumulation in cascaded pipelines.

Since pose features vary in reliability across denoising timesteps, a fixed injection strength is suboptimal. We therefore introduce a timestep-aware pose injection mechanism to adaptively modulate their contribution over the denoising trajectory without a predefined schedule.
To mitigate identity and appearance drift in long videos, we further propose persistent identity injection, which repeatedly injects complementary reference features into multiple RGB-stream blocks to improve motion fidelity and long-term visual consistency.


To promote research in this field, we introduce \textbf{FlowDanceSet}, a popularity-curated, high-resolution, in-the-wild dance video dataset collected from online videos. FlowDanceSet provides synchronized music and RGB videos together with 3D body motion, camera parameters, and 2D pose annotations, supporting both music-to-motion generation and music-driven dance video synthesis. Extensive experiments demonstrate that FlowDance achieves superior performance in both motion generation and video generation.

Our contributions are summarized as follows:
\begin{itemize}
    \item We propose FlowDance, a music-driven dance video generation framework with parallel streams for explicit motion modeling and reference-preserving video synthesis.
    \item We introduce timestep-aware pose injection for robust and adaptive structural guidance and persistent identity injection for long-term appearance consistency.
    \item We construct FlowDanceSet, a popularity-curated high-resolution in-the-wild dataset with music, RGB video, 3D motion, rich camera parameters, and 2D poses.
\end{itemize}

\flowdancewidefigure
    {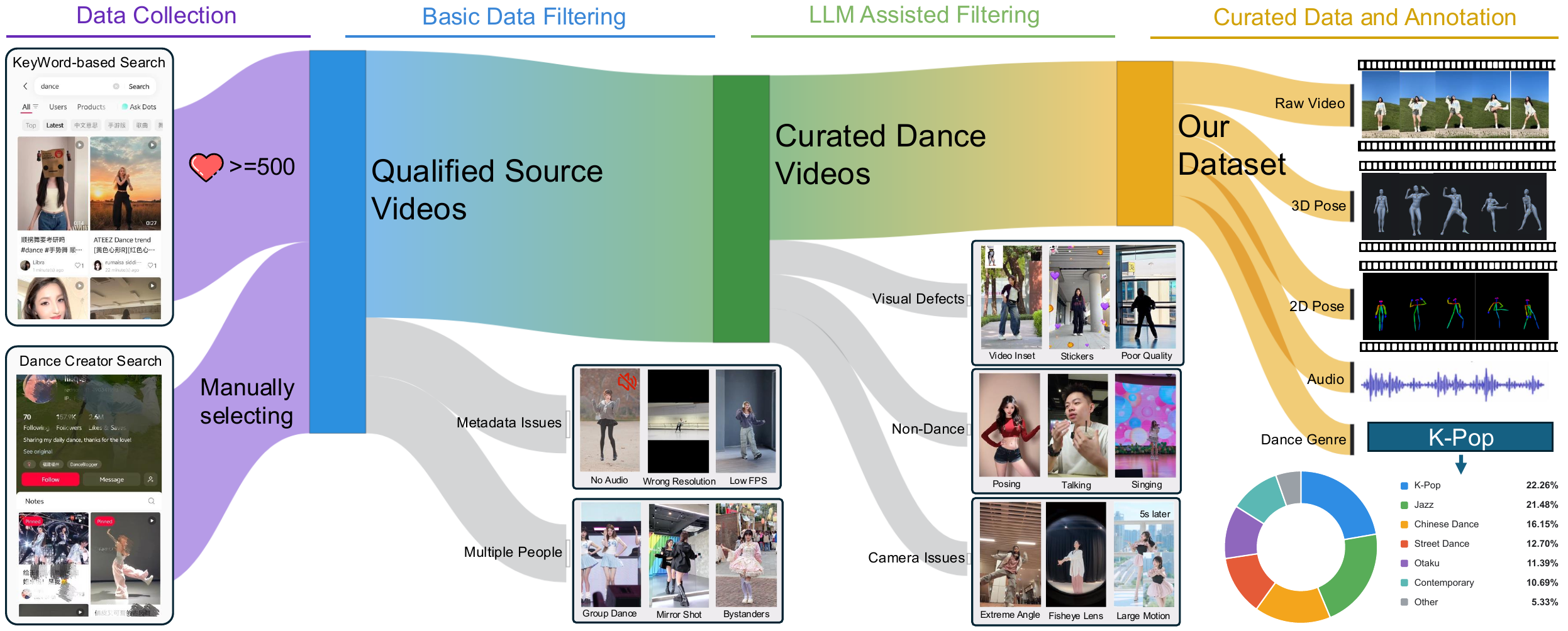}
    {FlowDanceSet construction pipeline. Collected dance videos are filtered for quality and annotated with synchronized RGB, audio, 3D motion, camera parameters, 2D poses, and dance labels.}
    {fig:dataset-pipeline}

\section{Related Work}
\subsection{Music-Driven Dance Generation}

Music-driven dance generation has predominantly targeted audio-conditioned motion rather than RGB video. 
Early systems composed reusable motion units or encoded choreography constraints~\cite{lee2019dancing,chen2021choreomaster}, and AIST++ established a popular benchmark with a cross-modal transformer baseline~\cite{li2021aist}. 
Later work improved long-horizon choreography and controllability with parametric transformers, choreographic memories, posture constraints, and coarse-to-fine diffusion architectures~\cite{li2021dancenet3d,siyao2022bailando,gao2022pcdance,li2024lodge}. 
Diffusion and multimodal formulations expanded editability and music-text control~\cite{tseng2022edge,qi2023diffdance,li2024lodge,gong2023tm2d}. 
Fine-grained data and contrastive or reward-guided objectives further improved full-body motion, diversity, and beat alignment~\cite{li2022finedance,bhattacharya2023danceanyway,wang2023e3d2}. 
Tokenized, popularity-oriented, large-scale, and editable formulations expanded control and data coverage~\cite{wang2024midget,luo2024popdg,zhang2025opendance,zhang2025danceeditor}. 
Most of these methods output joints, skeletons, or SMPL-family parameters~\cite{loper2015smpl,pavlakos2019smplx}, which remain intermediate representations for person-specific video synthesis.

Music-driven video methods either generate appearance-conditioned RGB videos more directly~\cite{wang2024dabfusion,dong2025musedance} or separate choreography and rendering through predicted skeletons and pose/body intermediates~\cite{ren2020selfsupervised,chen2025xdancer,wang2025choreomuse}. 
Direct formulations avoid an external motion-to-video stack but concentrate motion planning and photorealistic rendering in one generator; staged formulations simplify the rendering problem but depend on the accuracy and compatibility of the intermediate control.

\subsection{Pose-Guided Human Image Animation}

Pose-guided human image animation renders a reference person under an externally supplied motion signal. 
Earlier motion-transfer work used driving videos~\cite{zhou2019dance}; diffusion-based methods later combined reference-image conditioning with pose, expression, or disentangled controls to improve identity preservation and temporal consistency~\cite{karras2023dreampose,wang2023disco,chang2024magicpose,xu2023magicanimate,hu2023animate}. 
ControlNet provides a general mechanism for injecting spatial conditions into pretrained diffusion models~\cite{zhang2023controlnet}, while whole-body keypoints and richer 3D guidance strengthen structural control~\cite{yang2023dwpose,zhu2024champ,pang2024dreamdance}. Pose-text and hybrid-guidance systems further broaden motion, character, and background handling~\cite{qin2023dancingavatar,tan2025animatex,luo2025dreamactor}. 
Because these methods generally assume a driving pose sequence from a video, estimator, or upstream generator~\cite{xu2023magicanimate,hu2023animate,zhu2024champ,pang2024dreamdance}, they primarily address animation fidelity rather than music-conditioned choreography.

\flowdancewidefigure
    {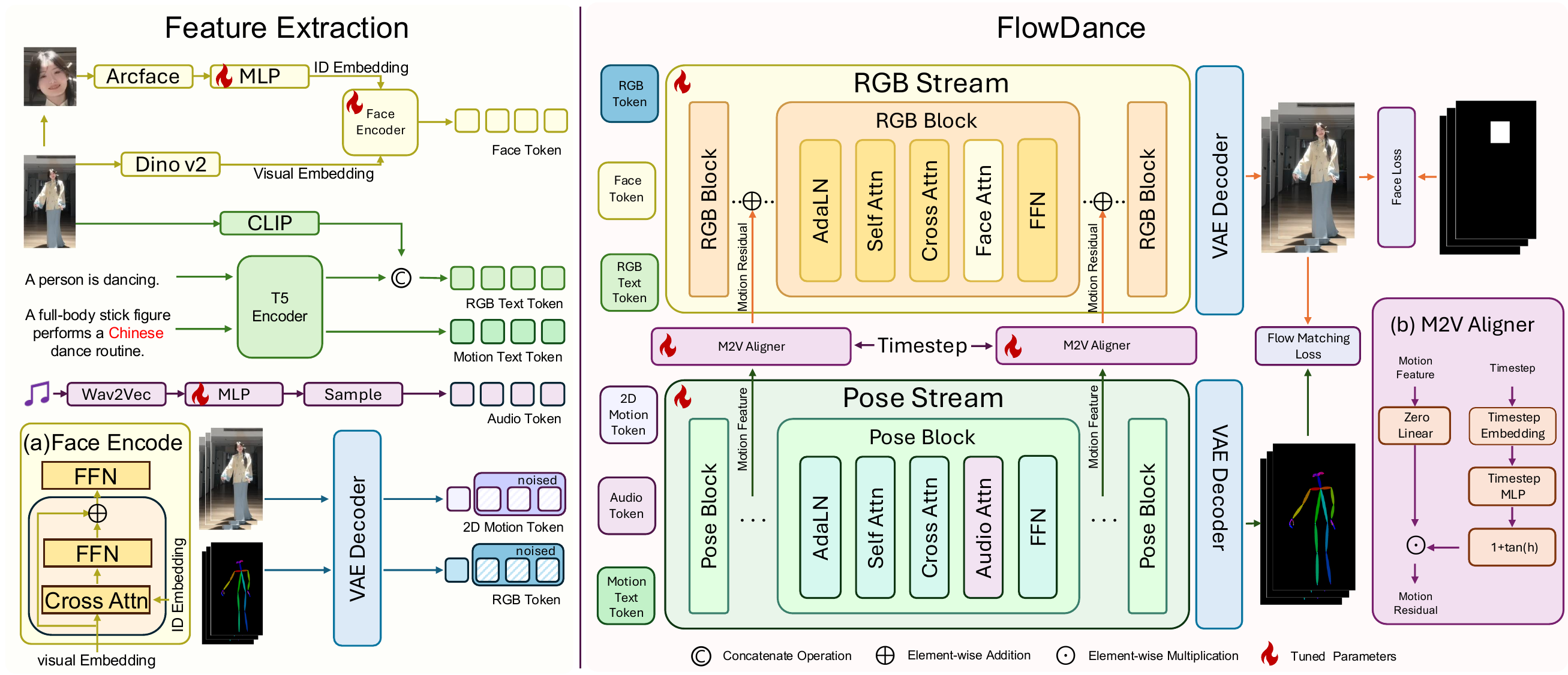}
    {Illustration of our FlowDance framework. \textbf{Left:} Feature extraction provides multimodal conditioning for the RGB and pose streams. \textbf{Right:} FlowDance comprises parallel yet coupled pose and RGB streams. The pose stream predicts music-aligned 2D motion, whose intermediate features guide RGB generation through timestep-aware M2V Aligners. Face-token injection and face-masked reconstruction further preserve identity. \textbf{Insets:} Details of (a) the Face Encoder and (b) the M2V Aligner.}
    {fig:method-pipeline}

\section{Dataset}
\textbf{Data collection.} Figure~\ref{fig:dataset-pipeline} summarizes the construction of FlowDanceSet. We select six dance categories with visually and stylistically distinct movement patterns: \textit{kpop}, \textit{jazz}, \textit{chinese\_dance}, \textit{otaku}, \textit{street\_dance}, and \textit{contemporary}. For each category, we use ChatGPT to construct a large set of related search keywords, including alternative names and platform-specific tags, and download only videos with at least 500 likes. This threshold provides an initial preference for videos that have been well received by viewers. However, popularity is also influenced by the performer's appearance, video editing techniques, and the choice of topics that cater to mainstream taste, rather than dance quality alone. We therefore complement keyword-based collection by manually selecting creators who consistently publish high-quality dance performances and downloading all of their videos.

\noindent\textbf{Data filtering.} We first apply a lightweight automatic filter, retaining videos that contain audio, have a resolution of $720\times1280$, have a frame rate of at least 16 FPS, and contain a single person according to MMPose. We then use Gemini 3 Flash for fine-grained visual screening based on three criteria: (1) \textit{minimal editing traces}, requiring a clear human contour without subtitles, stickers, picture-in-picture overlays, or other conspicuous editing elements; (2) \textit{dance validity}, requiring sustained dance movements rather than sexually suggestive posing, speaking, or singing; and (3) \textit{camera stability}, excluding strong camera movement, fisheye lenses, and extremely oblique viewpoints that impede motion assessment. Videos that fail any criterion are discarded.

\noindent\textbf{Data annotation.} Each case in the final dataset contains the original RGB video and audio, a dance-category label, 3D body-motion annotations with camera parameters, and a projected 2D pose video. During crawling, we also store the title, description, and tags of each source video. When these metadata explicitly identify one of the six dance categories, we directly use that label; for example, an \textit{otaku} tag provides an explicit category assignment. Gemini 3 Flash classifies the remaining videos, and samples that cannot be reliably assigned to the six predefined categories are labeled as \textit{other}. We use SAM-Body4D~\cite{gao2025sambody4d} to recover temporally consistent 3D body motion and the associated camera parameters from the full video. The 2D pose video is then obtained by projecting the recovered 3D motion with these camera parameters. We avoid directly using a 2D pose estimator because the fast movement and frequent self-occlusion in dance videos often cause missing leg keypoints and noticeable hand-keypoint drift. Before clip extraction, we annotate scene-transition timestamps in each full video and randomly reserve 100 source videos for evaluation. All extracted clips span five seconds (81 frames at 16~FPS) and avoid annotated transitions. We extract one clip from each held-out video to form a 100-case test set and exclude those source videos entirely from training. From the remaining videos, we extract clips with a two-second overlap. The RGB frames, 3D motion, and projected 2D annotations of every clip are aligned frame by frame. This process produces approximately 165K clips in total.

\section{Method}

Given a reference image $I_{\mathrm{ref}}$, a music clip $A$, and a dance-category label $c$, FlowDance synthesizes an RGB video $V=\{I_t\}_{t=1}^{T}$ of the reference person performing a category-consistent dance synchronized with the music. The model comprises two parallel latent video diffusion streams initialized from Wan2.2-TI2V-5B~\cite{wan2025}. The pose stream generates an auxiliary DWPose-style video $P=\{P_t\}_{t=1}^{T}$, while the RGB stream renders the reference person under its evolving structure. Both streams retain the image-to-video formulation of the pretrained model: the clean first-frame conditions are $I_{\mathrm{ref}}$ for the RGB stream and the pose extracted from $I_{\mathrm{ref}}$ for the pose stream, whereas the remaining frames are represented by noisy video latents. At diffusion timestep $t$, we denote the pose and RGB latents by $z_t^p$ and $z_t^r$, and their features at transformer block $l$ by $h_l^p$ and $h_l^r$, respectively. Rather than completing the pose video before RGB synthesis, the two streams predict their flow velocities in parallel, and the intermediate output of each pose block guides the corresponding RGB block. The pose stream thereby organizes the dance in a simplified visual space, while the RGB stream resolves appearance under continuous structural guidance.
Figure~\ref{fig:method-pipeline} overviews feature extraction, dual-stream denoising, and cross-stream interaction.

\subsection{Music-Conditioned Pose Video Generation}

The pose stream starts from the target person rather than from a canonical skeleton. We apply a whole-body pose estimator to $I_{\mathrm{ref}}$ and render the detected keypoints in the same DWPose representation used for the training videos~\cite{yang2023dwpose}. This pose image becomes $P_1$, the clean first-frame condition of the pose branch. The reference-derived initial skeleton establishes the generated dancer's body proportions, scale, and image-space location before motion unfolds, without requiring a separately parameterized 3D-to-2D projection.

The dance category and music describe different aspects of the trajectory. We instantiate the category prompt as ``A full-body stick figure performs a $c$ dance routine.'' and feed its text encoding to the pretrained cross-attention layers, providing a clip-level description of dance style. The music, by contrast, enters every pose block as a temporally aligned signal. We first resample the waveform to 16 kHz and encode it with a frozen Wav2Vec2 model~\cite{baevski2020wav2vec}. Features from its transformer layers are retained and linearly interpolated to the temporal length of the video latent.

Audio injection is local in time. For a pose token belonging to latent frame $f$, the keys and values are drawn from the audio features of a five-frame sliding window

\[
\mathcal{W}(f)=\{\max(1,f-2),\ldots,\min(F,f+2)\},
\]

where $F$ is the number of latent frames and indices outside the sequence are clamped to its boundary. Consequently, each pose frame can follow the nearby musical context more selectively without attending indiscriminately to the full clip. The resulting cross-attention output is passed through a zero-initialized projection before it is added to the pose features, thereby preserving the pretrained video computation unchanged at the start of training.

\subsection{Reference-Preserving RGB Video Synthesis}

The RGB stream receives global reference information through its main cross-attention context. We encode the fixed prompt ``A person is dancing.'' with the pretrained T5 text encoder and concatenate its text features with full-image features extracted from $I_{\mathrm{ref}}$ by a frozen CLIP image encoder~\cite{radford2021clip}. The resulting RGB conditioning tokens provide both a generic action description and reference-specific cues for appearance and scene content.

Structural information arrives from the pose stream at the transformer-block level. After pose block $l$, we project $h_l^p$ and add it to the input of the corresponding RGB block. Pose features do not have the same reliability at all noise levels, so the current timestep embedding $e_t$ also controls the transferred residual:

\[
\widetilde{h}_l^r
=h_l^r+W_l(h_l^p)\odot\left[1+\tanh\left(G_l(e_t)\right)\right],
\]

where $W_l$ projects pose features, $G_l$ is a two-layer timestep projection, and its output is broadcast over the video tokens. The RGB block then processes $\widetilde{h}_l^r$. We zero initialize $W_l$ and the final layer of $G_l$, allowing the interaction to grow from the original RGB backbone during training rather than perturbing it at initialization.

A compact ArcFace embedding~\cite{deng2019arcface} identifies the reference person but provides limited visual detail for faithful rendering. Therefore, we use a Perceiver-style Face Encoder to combine the normalized ArcFace embedding with DINOv2 features from $I_{\mathrm{ref}}$~\cite{oquab2024dinov2}, producing four complementary face tokens. Dedicated face-attention adapters inject these tokens into selected RGB blocks, while zero-initialized output projections preserve the pretrained computation at initialization.

For the RGB stream, flow matching constructs $z_t^r=(1-\sigma_t)z_0^r+\sigma_t\epsilon$ with target velocity $v_t^r=\epsilon-z_0^r$. Excluding the conditioned first frame, the primary RGB objective is

\[
\mathcal{L}_{\mathrm{rgb}}
=\mathop{\mathrm{E}}_{z_0^r,\epsilon,t}
\left[w(\sigma_t)\left\|\widehat{v}_t^r-v_t^r\right\|_2^2\right],
\]

where $w(\sigma_t)$ denotes the timestep-dependent loss weighting. To place additional supervision on facial reconstruction, we estimate the clean latent as $\widehat{z}_0^r=z_t^r-\sigma_t\widehat{v}_t^r$. With a face mask $M_{\mathrm{face}}$ downsampled to latent resolution, we define

\[
\mathcal{L}_{\mathrm{face}}=
\frac{\sum M_{\mathrm{face}}\odot(\widehat{z}_0^r-z_0^r)^2}
{C\sum M_{\mathrm{face}}+\varepsilon},
\]

where $C$ is the latent channel count. The conditioned first frame is excluded from the sum. This masked term is added to the RGB flow-matching objective and does not alter the reconstruction weighting outside the detected face.

\subsection{Training and Inference}

Training proceeds in two stages because the pose and RGB streams solve problems of different complexity. The first stage adapts the copied video backbone to pose videos using only the pose flow-matching objective. It begins with a low-resolution warm-up and then continues at full resolution. During this stage, only the pose stream and its audio-injection modules are optimized. Once the pose stream has learned the music-conditioned motion space, its parameters are frozen. The second stage trains the RGB stream, the pose-to-RGB projections, and the reference and face conditioning modules with $\mathcal{L}=\mathcal{L}_{\mathrm{rgb}}+\lambda_{\mathrm{face}}\mathcal{L}_{\mathrm{face}}$. The pretrained Wav2Vec2, T5, CLIP, DINOv2, and ArcFace encoders remain frozen in both stages.

Music longer than the training clip is handled with overlap-window inference. Specifically, we divide the audio and conditioning sequence into fixed-length windows and generate them sequentially with a short temporal overlap. Frames produced in the overlap of the previous window condition the next one, after which the two predictions are linearly blended over their shared interval, thereby reducing visible discontinuities across adjacent windows. The same model can thus extend generation without increasing the temporal memory used by a single denoising run.

\flowdancewidefigure
    {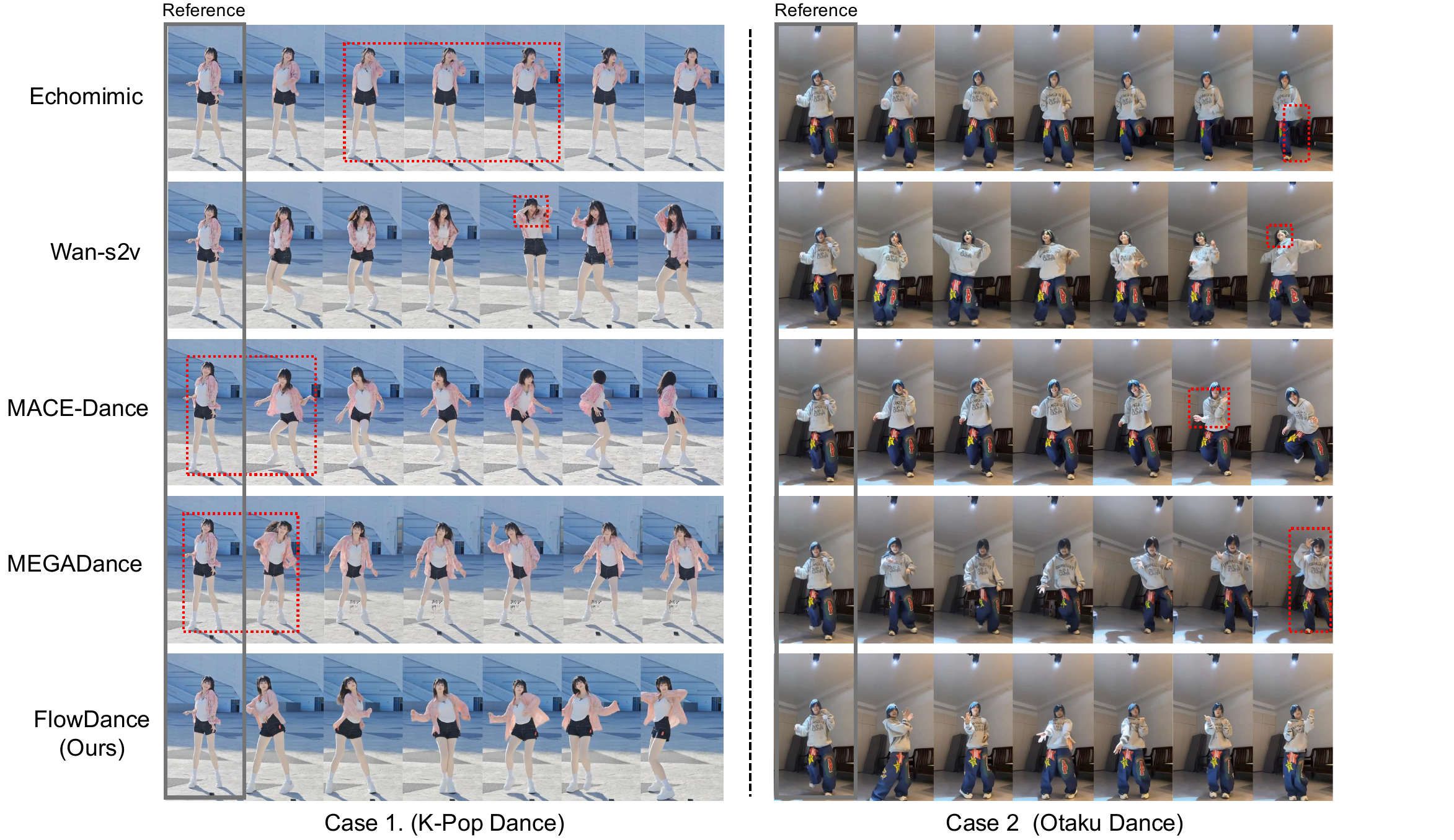}
    {Qualitative comparison on unseen K-pop and otaku examples. Red boxes mark weak motion, deformation, identity drift, or implausible structure. FlowDance preserves appearance while producing larger, coherent movements.}
    {fig:qualitative}

\section{Experiments}

\subsection{Implementation Details}

We initialize both streams from Wan2.2-TI2V-5B. Audio is resampled to 16~kHz, and each training sample pairs a five-second, 81-frame video clip at 16~FPS with its temporally aligned audio. The Face Encoder produces four face tokens, which are injected into RGB blocks 4, 12, 20, and 28, and we set $\lambda_{\mathrm{face}}=0.3$. Training is performed on eight NVIDIA A100 GPUs. In Stage~1, we train the pose stream for 30K warm-up steps at $352\times640$ and then for 6K steps at $704\times1280$; Stage~2 is trained for 20K steps at $704\times1280$. We use AdamW in both stages with a learning rate of $2\times10^{-5}$, a weight decay of 0.03, and 100 learning-rate warm-up steps followed by a constant schedule. The per-GPU batch size is 1, with gradient accumulation over four iterations, giving an effective batch size of 32. At inference, we use the UniPC sampler with 50 sampling steps and a classifier-free guidance scale of 6.0.

\subsection{Evaluation Protocol}

\textbf{Baselines.} We compare against two cascaded motion-to-video pipelines and two end-to-end audio-to-video systems. For the cascaded setting, we use the Motion Expert of MACE-Dance~\cite{yang2025macedance}\footnote{We do not evaluate the MACE-Dance Appearance Expert since its released implementation cannot yield stable inference in our tests.} and MEGADance~\cite{yang2025megadance} to generate 3D motion. To render a pose video for a target identity, we use SAM-Body4D to recover the body shape, image-space location, and camera parameters from the reference image, keep these quantities fixed, and animate the recovered body with the generated motion. The resulting pose sequence drives One-to-All Animation~\cite{shi2026onetoall}, whose alignment-robust design is intended to reduce reference--pose mismatch. The end-to-end baselines are Wan-S2V~\cite{gao2025wans2v} and EchoMimic~\cite{chen2024echomimic}.
We fine-tune Wan-S2V and MEGADance on the FlowDanceSet training split using their released training code; MACE-Dance and EchoMimic release no training code, so we evaluate them zero-shot with their official checkpoints.
We evaluate all methods on the 100-case, source-video-held-out test set described above.


\textbf{Metrics.} We measure distribution-level video quality using Fr'echet Video Distance (FVD)~\cite{ge2024contentbias}. We additionally report six VBench dimensions~\cite{huang2024vbench}: imaging quality (IQ), aesthetic quality (AQ), subject consistency (SC), background consistency (BC), motion smoothness (MS), and dynamic degree (DD). For direct comparability, motion evaluation is performed on the final RGB output of every method rather than on an internal motion representation. Specifically, SAM-Body4D extracts a 3D sequence from each generated video, after which evaluation is restricted to the same 24 body joints for all methods. Following common music-to-dance evaluation~\cite{li2021aist,yang2025macedance}, we report kinetic and geometric Fr'echet distances (FID$_k$/FID$_g$), their corresponding diversity scores (Div$_k$/Div$_g$), and Beat Alignment Score (BAS).

\begin{table*}[t]
    \centering
    \caption{Quantitative comparison on the test set. Arrows indicate preferred directions. The best and second-best generated results in each column are highlighted in \textbf{bold} and \underline{underlined}, respectively. GT is an unranked reference.}

    \label{tab:quantitative}
    \scriptsize
    \setlength{\tabcolsep}{1.25pt}
    \resizebox{\textwidth}{!}{%
        \begin{tabular}{@{}lcccccccccccccccc@{}}
            \toprule
            & \multicolumn{7}{c}{Video quality} & \multicolumn{5}{c}{Motion quality} & \multicolumn{4}{c}{User preference (\%)} \\
            \cmidrule(lr){2-8}\cmidrule(lr){9-13}\cmidrule(l){14-17}
            Method & FVD$\downarrow$ & IQ$\uparrow$ & AQ$\uparrow$ & SC$\uparrow$ & BC$\uparrow$ & MS$\uparrow$ & DD$\uparrow$ & FID$_k\downarrow$ & FID$_g\downarrow$ & Div$_k\uparrow$ & Div$_g\uparrow$ & BAS$\uparrow$ & VQ$\uparrow$ & ID$\uparrow$ & MN$\uparrow$ & M--A$\uparrow$ \\
            \midrule
            GT & -- & 0.700 & 0.539 & 0.926 & 0.931 & 0.983 & 0.991 & -- & -- & 6.357 & 7.329 & 0.2331 & -- & -- & -- & -- \\
            MACE-Dance~\cite{yang2025macedance} & 662.4355 & 0.689 & 0.522 & 0.948 & \underline{0.952} & \underline{0.991} & 0.347 & 51.466 & 92.749 & 6.673 & 8.267 & 0.2176 & 5.4 & 3.6 & 5.0 & 5.9 \\
            MEGADance~\cite{yang2025megadance} & 409.5303 & 0.668 & 0.524 & 0.940 & 0.941 & 0.988 & 0.906 & \textbf{9.271} & 13.243 & 6.132 & 7.444 & 0.2160 & 3.6 & 3.8 & 5.3 & \underline{13.8} \\
            EchoMimic~\cite{chen2024echomimic} & 571.0114 & \underline{0.693} & 0.510 & \textbf{0.954} & \textbf{0.954} & \underline{0.991} & 0.008 & 38.699 & 1087.053 & 3.757 & \textbf{27.625} & 0.2088 & 11.0 & \underline{20.8} & 1.0 & 0.9 \\
            Wan-S2V~\cite{gao2025wans2v} & \underline{328.0243} & 0.671 & \textbf{0.537} & 0.924 & 0.936 & 0.980 & \underline{0.979} & 16.069 & \underline{12.8056} & \textbf{7.386} & 7.829 & \underline{0.2411} & \underline{15.0} & 12.3 & \underline{24.5} & 9.9 \\
            \midrule
            FlowDance (ours) & \textbf{243.3741} & \textbf{0.694} & \underline{0.526} & \underline{0.951} & \underline{0.952} & \textbf{0.993} & \textbf{0.983} & \underline{10.884} & \textbf{7.501} & \underline{6.685} & \underline{8.304} & \textbf{0.2450} & \textbf{65.0} & \textbf{59.6} & \textbf{64.3} & \textbf{69.6} \\
            \bottomrule
        \end{tabular}%
    }
\end{table*}

\begin{table}[t]
    \centering
    \caption{Component ablation. \textbf{Bold} and \underline{underline} denote the best and second-best results.}
    \label{tab:ablation}
    \footnotesize
    \setlength{\tabcolsep}{2.2pt}
    \begin{tabular}{@{}lccccc@{}}
        \toprule
        Variant & FVD$\downarrow$ & SC$\uparrow$ & MS$\uparrow$ & FID$_g\downarrow$ & BAS$\uparrow$ \\
        \midrule
        w/o pose stream & 387.06 & 0.876 & 0.962 & 34.166 & 0.2081 \\
        fixed pose injection & 278.63 & \underline{0.946} & 0.983 & 8.943 & 0.2321 \\
        ArcFace only & \underline{249.86} & 0.937 & 0.990 & \underline{7.616} & \underline{0.2436} \\
        w/o face reconstruction loss & 254.45 & 0.931 & \underline{0.991} & 7.998 & 0.2402 \\
        Full model (ours) & \textbf{243.37} & \textbf{0.951} & \textbf{0.993} & \textbf{7.501} & \textbf{0.2450} \\
        \bottomrule
    \end{tabular}
\end{table}

\subsection{Quantitative Results}

Table~\ref{tab:quantitative} shows that FlowDance obtains the lowest FVD (243.3741), 25.8\% below Wan-S2V, and the best imaging quality, motion smoothness, and dynamic degree among generated methods. EchoMimic has slightly higher subject and background consistency, but its DD of 0.008 indicates an almost static solution; its large FID$_g$ and Div$_g$ show that structural artifacts can inflate feature dispersion. FlowDance instead combines SC of 0.951 with DD of 0.983.

Recovered-motion metrics separate geometric plausibility from audio responsiveness. MEGADance obtains the lowest FID$_k$ (9.271) through direct 3D-motion optimization, while FlowDance ranks second (10.884) and achieves the best FID$_g$ (7.501) and BAS (0.2450). Wan-S2V remains competitive elsewhere, but its FID$_g$ is 70.7\% higher. These results support an explicit low-complexity motion stream coupled to RGB synthesis rather than an isolated cascaded renderer.

\subsection{Qualitative Results}

Figure~\ref{fig:qualitative} compares two visually distinct identities and dance categories. EchoMimic largely preserves the source composition but produces limited body displacement, consistent with its near-zero DD. Wan-S2V generates larger motion and strong image quality, yet exhibits local face and limb distortions in challenging frames. The cascaded baselines produce recognizable choreography, but errors introduced by 3D motion projection and subsequent rendering appear as abrupt changes in body scale, local deformation, and weaker identity fidelity. FlowDance follows the music with clear full-body articulation while maintaining clothing, facial appearance, and scene structure across the sampled frames. The benefit is particularly visible in Case~2, where the loose clothing and large limb motion make reference--pose mismatch difficult for a cascaded pipeline. Figure~\ref{fig:long-video} further shows a 12-second generated sequence. This example indicates that repeated identity injection and overlap conditioning can limit cross-window appearance drift.

\flowdancecolumnfigure
    {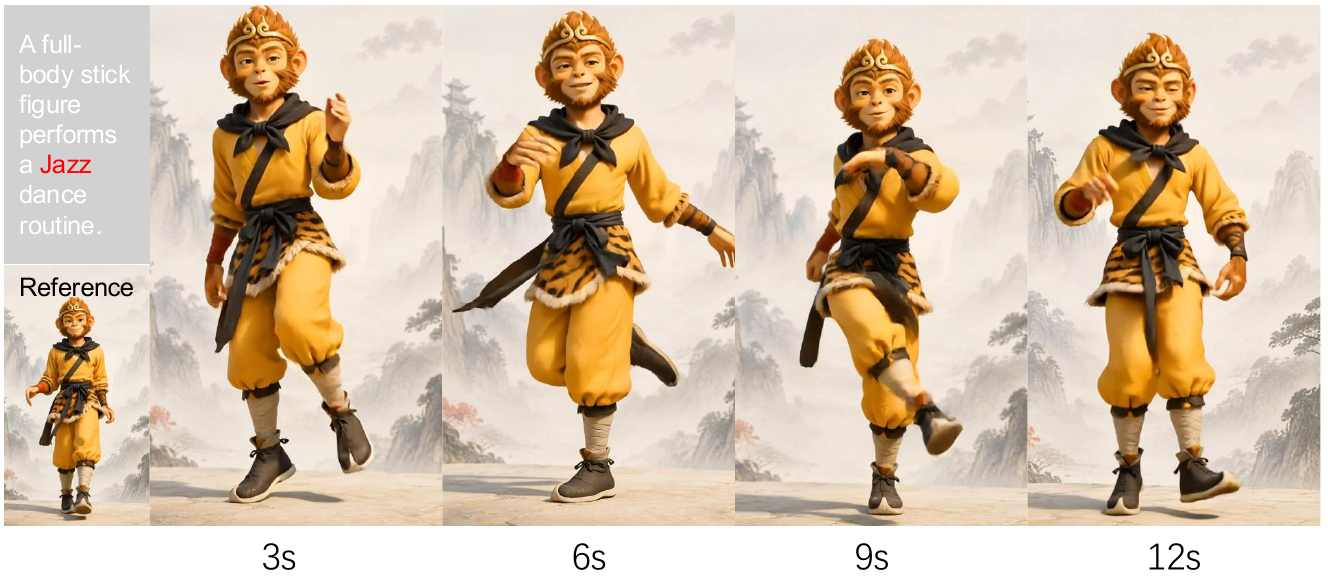}
    {A 12-second sequence generated by overlap-window inference. Identity, costume, and scene remain stable as the pose evolves.}
    {fig:long-video}

\subsection{User Study}

In a blinded user study, 40 participants evaluate the same 20 random test cases. Five anonymized outputs per case are randomized under identical audio, reference, resolution, and playback settings. Selecting the best visual quality (VQ), identity fidelity (ID), motion naturalness (MN), and music--motion alignment (M--A) yields 800 selections per criterion. FlowDance leads all four, with 65.0\% VQ, 59.6\% ID, 64.3\% MN, and 69.6\% M--A preference.

\subsection{Ablation Study}

We isolate four design choices: removing the pose stream adds audio cross-attention directly to RGB; fixed residual injection tests the M2V Aligner; ArcFace-only conditioning removes the DINOv2--Face Encoder pathway; and removing $\mathcal{L}_{\mathrm{face}}$ tests local facial supervision. All variants share the full model's data, backbone, optimization, and inference settings.


Table~\ref{tab:ablation} shows that removing the pose stream causes the largest degradation, increasing FVD from 243.37 to 387.06 and FID$_g$ from 7.501 to 34.166 while reducing BAS from 0.2450 to 0.2081. Fixed injection also underperforms, confirming the benefit of timestep awareness. ArcFace-only conditioning or removing face reconstruction degrades FVD and subject consistency; the full model leads all five metrics.

\section{Conclusion}

We introduced FlowDance, a parallel pose-video generation framework that models choreography in a simplified pose stream while rendering the target person in an RGB stream. Timestep-aware pose injection couples motion and appearance generation, while persistent identity injection improves appearance consistency over extended sequences. Together with FlowDanceSet, experiments demonstrate strong motion quality, video quality, and music alignment. Future work will explore adapting the parallel two-stream architecture for streaming, real-time dance video generation.

\bibliography{aaai2027}


\end{document}